\pdfoutput=1

\documentclass[11pt]{article}

\usepackage{acl}

\usepackage{times}
\usepackage{latexsym}

\usepackage[T1]{fontenc}

\usepackage[utf8]{inputenc}

\usepackage{microtype}

\usepackage{inconsolata}

\usepackage{algorithm}
\usepackage{algpseudocode}
\usepackage{amsmath}
\usepackage{tabularx}
\usepackage{booktabs}
\usepackage{subcaption}
\usepackage{multirow}
\usepackage{amsfonts,amssymb,bbding,pifont,makecell,bm,diagbox}

\usepackage{newfloat}
\usepackage{listings}
\usepackage{graphicx}
\usepackage{xurl}
\usepackage{xparse}
\usepackage{array}
\usepackage{stfloats}

\usepackage{helvet}
\usepackage{colortbl}  
\usepackage{xcolor}
\usepackage[normalem]{ulem}
\usepackage{ragged2e}
\usepackage{CJKutf8}

\usepackage{array}
\usepackage{ragged2e}
\newcolumntype{C}[1]{>{\Centering\arraybackslash}m{#1}}

\title{Listen Again and Choose the Right Answer: A New Paradigm for Automatic Speech Recognition with Large Language Models}

\author{Yuchen Hu$^1$, Chen Chen$^1$, Chengwei Qin$^1$, Qiushi Zhu$^{2}$, Eng Siong Chng$^1$, Ruizhe Li$^3$\thanks{Corresponding author.} \\
$^1$Nanyang Technological University, Singapore\\ $^2$University of Science and Technology of China, China \quad $^3$University of Aberdeen, UK \\
\small{\texttt{yuchen005@e.ntu.edu.sg, ruizhe.li@abdn.ac.uk}}
}

\begin{document}
\maketitle
\begin{abstract}
Recent advances in large language models (LLMs) have promoted generative error correction (GER) for automatic speech recognition (ASR), which aims to predict the ground-truth transcription from the decoded N-best hypotheses.
Thanks to the strong language generation ability of LLMs and rich information in the N-best list, GER shows great effectiveness in enhancing ASR results.
However, it still suffers from two limitations: 1) LLMs are unaware of the source speech during GER, which may lead to results that are grammatically correct but violate the source speech content, 2) N-best hypotheses usually only vary in a few tokens, making it redundant to send all of them for GER, which could confuse LLM about which tokens to focus on and thus lead to increased miscorrection.
In this paper, we propose ClozeGER, a new paradigm for ASR generative error correction.
First, we introduce a multimodal LLM (\emph{i.e.}, SpeechGPT) to receive source speech as extra input to improve the fidelity of correction output.
Then, we reformat GER as a cloze test with logits calibration to remove the input information redundancy and simplify GER with clear instructions.
Experiments show that ClozeGER achieves a new breakthrough over vanilla GER on 9 popular ASR datasets.
\end{abstract}


\section{Introduction}
\label{sec:intro}
Recent advances in large language models (LLMs) have attracted a surge of research interest thanks to their remarkable language generation and reasoning ability~\citep{chatgpt,gpt4,touvron2023llama,touvron2023llama2}, which achieve a wide range of success on natural language processing (NLP) tasks~\citep{brown2020language,wei2022emergent,ouyang2022training}.
Powered by LLMs, latest work~\citep{chen2023hp,chen2023generative,chen2024its,hu2024large,hu2024gentranslate} proposes a generative error correction~\citep{yang2023generative} (GER) benchmark\footnote{\url{https://github.com/Hypotheses-Paradise/Hypo2Trans}} for automatic speech recognition (ASR), and they release a HyPoradise dataset\footnote{\url{https://huggingface.co/datasets/PeacefulData/HP-v0}} that contains over 332K pairs of decoded N-best hypotheses and ground-truth transcription in various ASR domains.
It has shown great effectiveness in learning the mapping from hypotheses to transcription by parameter-efficient LLM finetuning~\citep{hu2021lora}, which significantly enhances the ASR result and outperforms typical LM rescoring methods~\citep{mikolov2010recurrent}.


\begin{figure}[t]
\begin{center}
\includegraphics[width=0.97\columnwidth]{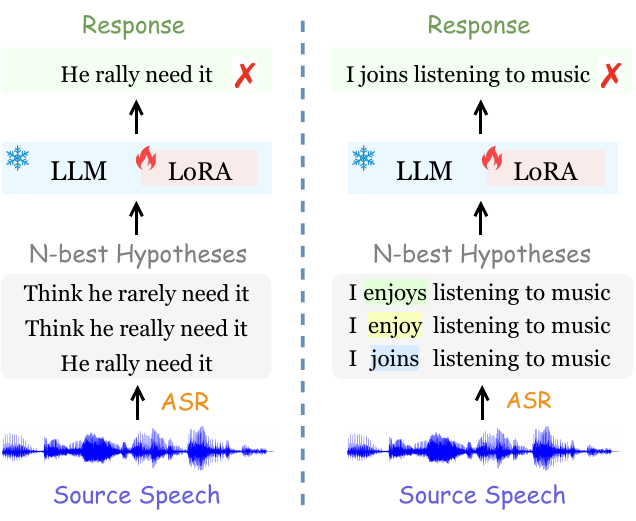}
\end{center}
\vspace{-0.2cm}
\caption{Two limitations of generative error correction~\citep{chen2023hp}. \textbf{Left: violate source speech}, LLM removes the word ``Think'' in first two hypotheses as it rarely appears at the beginning of a sentence and followed by a subject according to grammar, but this actually happens in the source speech.
\textbf{Right: information redundancy in N-best hypotheses input}, there is only one difference between N-best candidates, making it redundant to send all of them for GER, which confuses LLM about which tokens to focus on for correction.
}
\vspace{-0.3cm}
\label{fig1}
\end{figure}

However, GER paradigm is also observed to suffer from two limitations.
First, LLMs are unaware of the source speech during GER process, which could lead to results that do not match the source speech content.
For example, as shown in Fig.~\ref{fig1} (left), the source speech reads the word ``Think'' at the beginning and followed by ``he'', which is correctly recognized by the 1-best hypothesis.
Then during the GER process, LLM removes the word ``Think'', as this structure of verb plus noun at the beginning of a sentence is not rigorous according to grammar.
However, this is not expected as it violates the source speech content.
Second, we observe that N-best hypotheses usually only vary in a few tokens.
For example, as shown in Fig.~\ref{fig1} (right), all the tokens in candidates are the same except ``enjoys''/``enjoy''/``joins''.
In this case, it would be information redundant to leverage all of the hypotheses for predicting the ground-truth transcription, which could confuse the LLMs about which tokens to focus on for correction and thus lead to sub-optimal GER performance. 

Motivated by the above observations, we propose ClozeGER, a new paradigm for ASR generative error correction.
First, we introduce a popular multimodal LLM, SpeechGPT~\citep{zhang2023speechgpt}, to receive source speech as an extra input to the GER paradigm.
With the powerful cross-modal ability of SpeechGPT, we can now constrain GER to comply with the source speech while correcting the errors in decoded hypotheses.
Then, in order to remove the input information redundancy, we reformat it as a cloze test (\emph{i.e.}, a special multiple-choice question) with logits calibration~\citep{kumar2022answer,wang2023large}, where the identical parts across N-best hypotheses are set as the context and the varying parts are set as blanks (each with several options provided).
With such clear instructions for error correction, it would be easier for LLMs to perform context reasoning and choose the right answer for each blank rather than predicting the entire sentence from redundant N-best inputs\footnote{Think if we humans are asked to do GER, which option is easier and efficient, cloze or entire sentence prediction?}.
Finally, we add a simple post-processing stage to correct the errors in cloze context (\emph{i.e.}, identical parts across N-best list) to further improve the correction result.

Our contributions are summarized as follows:

\begin{itemize}
    \item We propose ClozeGER, a new paradigm based on multimodal LLM for ASR generative error correction, which receives both source speech and the decoded N-best hypotheses as input to predict the ground-truth transcription.
    \item We further reformat the generative error correction as a cloze test with logits calibration to remove the information redundancy in N-best hypotheses input and simplify the GER paradigm with clear instructions.
    \item Experiment evidence shows that our proposed ClozeGER achieves a new breakthrough over vanilla GER on 9 popular ASR datasets.
\end{itemize}

\section{Related Work}
\label{sec:related_work}

\noindent\textbf{Large Language Models.}
There is recently a surge of research interests in Transformer-based LLMs, such as ChatGPT~\citep{chatgpt}, GPT-4~\citep{gpt4} and LLaMA~\citep{touvron2023llama,touvron2023llama2}.
Benefiting from the huge model size and abundant training data, LLMs can well understand the linguistic structures and semantic meanings behind textual data, which shows remarkable performance on a wide range of NLP tasks~\citep{brown2020language,wei2022emergent,ouyang2022training}.
More recently, researchers have started to explore the potential of LLMs on multimodal tasks by incorporating other modalities into LLMs~\citep{liu2023visual,li2023blip,chen2023x,zhang2023video,zhang2023llama,gao2023llama,fathullah2023prompting}.
Among them, SpeechGPT~\citep{zhang2023speechgpt} is one of the most popular multimodal LLMs that represent speech and text using a unified tokenizer, which enables us to add source speech into the original N-best hypotheses input of the GER paradigm.

\vspace{0.1cm}
\noindent\textbf{LM Rescoring and ASR Generative Error Correction.}
LM rescoring has been widely used in ASR decoding to rerank the N-best hypotheses and yield better 1-best recognition result~\citep{arisoy2015bidirectional,shin2019effective,mikolov2010recurrent}.
Furthermore, to make full use of all candidatures, recent works employ the entire N-best list for error correction~\citep{leng2021fastcorrect2,ma2023n}.
Powered by LLMs, latest work proposes a generative error correction (GER) benchmark~\citep{chen2023hp} to predict the ground-truth transcription from ASR N-best hypotheses and achieves remarkable performance.
This work serves as an extension of GER to resolve the existing limitations.

\vspace{0.1cm}
\noindent\textbf{Cloze Test with LLMs.}
As a special format of multiple-choice questions (MCQ), the cloze test provides a context with some blanks, where each blank is provided with several options for selection.
Recently, cloze test and MCQ are widely adopted in LLM-centric scenarios~\citep{chiang2023vicuna,zheng2023judging}, as well as numerous LM benchmarks including MMLU~\citep{hendrycks2020measuring}, AGIEval~\citep{zhong2023agieval}, and C-Eval~\citep{huang2023c}.
However, recent works observe that LLMs-based cloze test is vulnerable to option position changes due to their inherent ``selection bias''~\citep{kumar2022answer,wang2023large,pezeshkpour2023large}.
In this work, we reformat the GER paradigm as a cloze test for simplification, as well as introduce a logits calibration method to remove the existing selection bias and make LLM a robust cloze solver.

\section{Methodology}
\label{sec:method}
In this section, we present our proposed ClozeGER paradigm in detail.
We first introduce the preliminary knowledge of GER in \S\ref{ssec:pre_ger}, and then we investigate to introduce source speech to GER paradigm with multimodal LLM (\S\ref{ssec:ger_with_speech}).
Finally, we present the new task format of ClozeGER in \S\ref{ssec:clozeger}.

\begin{figure*}[t]
\begin{center}
\includegraphics[width=0.98\textwidth]{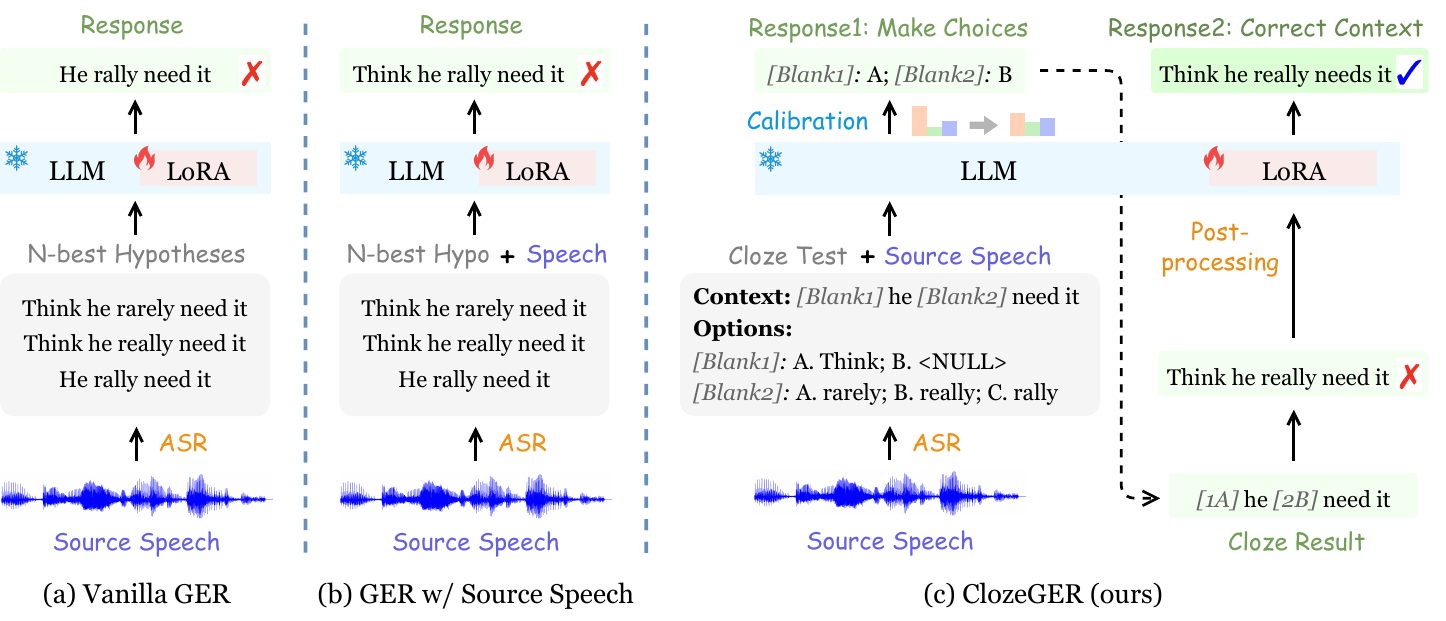}
\end{center}
\vspace{-0.2cm}
\caption{Frameworks of (a) vanilla GER that employs N-best hypotheses to predict ground-truth transcription, (b) GER with source speech as extra input to improve the fidelity of correction output, (c) our ClozeGER that reformats GER as a cloze test with logits calibration, followed by a post-processing stage to further correct the cloze context.}
\label{fig2}
\vspace{-0.2cm}
\end{figure*}

\subsection{Preliminary: Generative Error Correction}
\label{ssec:pre_ger}
We follow the original generative error correction benchmark~\citep{chen2023hp} as shown in Fig.~\ref{fig2} (a).
Given an input speech $X$, the pre-trained ASR model first transcribe it into $N$-best hypotheses $\mathcal{Y}_N = \{Y_1,Y_2,\cdots,Y_N\}$ by beam search decoding.
The goal of GER is to learn a hypotheses-to-transcription (H2T) mapping $\mathcal{M}_\text{H2T}$ that predicts the transcription $Y$ based on $N$-best list $\mathcal{Y}_N$:
\begin{equation}
\label{eq1}
\begin{aligned}
    Y &= \mathcal{M}_\text{H2T} (\mathcal{Y}_N),
\end{aligned}
\end{equation}
Given the ground-truth transcription $Y^*$, we can finetune the LLM to learn $\mathcal{M}_\text{H2T}$ in an auto-regressive manner, where the cross-entropy loss $\mathcal{L}_\text{H2T}$ is formulated as:
\begin{equation}
\label{eq2}
\begin{aligned}
    \mathcal{L}_\text{H2T} &= \sum_{t=1}^T -\log \mathcal{P}_{\theta} (y_t^* | y_{t-1}^*, \cdots, y_1^*, \mathcal{Y}_N),
\end{aligned}
\end{equation}
where $y_t^*$ is the $t$-th token of $Y^*$, and $\theta$ denotes the learnable parameters in LLM (\emph{i.e.}, LoRA).

\subsection{GER with Source Speech}
\label{ssec:ger_with_speech}
In order to prevent GER from violating the content of source speech, we incorporate it as an extra input into LLM to improve output fidelity as shown in Fig.~\ref{fig2} (b).
So that Eq.(\ref{eq1}) should be rewritten as:
\begin{equation}
\label{eq3}
\begin{aligned}
    Y &= \mathcal{M}_\text{H2T} (\mathcal{Y}_N, X),
\end{aligned}
\end{equation}
where the N-best hypotheses and source speech are concatenated using the following instructions:

\emph{"Below is a speech and its candidate transcriptions from a speech recognition system.
Please listen to the speech and revise the candidate transcriptions to generate better final recognition results.
\#\#\# Speech:\{speech units\}.
\#\#\# Candidates:\{N-best hypotheses\}.
\#\#\# Response: "}

To jointly process text and speech, we leverage the popular multimodal LLM, SpeechGPT\footnote{\url{https://huggingface.co/fnlp/SpeechGPT-7B-cm}}~\citep{zhang2023speechgpt}, to replace the LLaMA in original GER benchmark.
Notably, SpeechGPT is developed by discretizing speech into 1,000 HuBERT units and adding them to LLaMA-7b\footnote{\url{https://huggingface.co/yahma/llama-7b-hf}} tokenizer, and it then finetunes LLaMA-7b to learn cross-modality mapping.
With such multimodal ability, we can enable GER to comply with source speech.

\subsection{ClozeGER}
\label{ssec:clozeger}

\subsubsection{Cloze Format}
\label{sssec:cloze_format}
Since N-best hypotheses usually only vary in a few tokens, it would be information redundant to send all of them for GER, which could confuse the LLM about which tokens to focus on and thus lead to increased miscorrection.
To this end, we simplify the GER paradigm as a cloze test as shown in Fig.~\ref{fig2} (c).
Specifically, we set the identical parts across N-best hypotheses as the context and the varying parts as blanks, where each blank is provided with several options.
In addition, we also insert a null token `<NULL>' to align the N-best candidates.
We design an instruction-following cloze template:

\emph{"Below is a speech and its candidate transcriptions from a speech recognition system. The candidates are formatted as a cloze test, where the blanks to fill are indicated by `\textcolor[HTML]{666666}{[Blank1]}', `\textcolor[HTML]{666666}{[Blank2]}', etc. 
Each blank is provided with several options indicated by ID letters `A', `B', `C', etc., where `<NULL>' denotes the null token. 
To generate a better final recognition result, please listen to the speech and write an answer choice for each blank.
\#\#\# Speech:\{speech units\}. 
\#\#\# Cloze test: \textcolor[HTML]{666666}{[Blank1]} he \textcolor[HTML]{666666}{[Blank2]} need it.
\#\#\# Options: \textcolor[HTML]{666666}{[Blank1]}: A. Think; B. <NULL>. \textcolor[HTML]{666666}{[Blank2]}: A. rarely; B. really; C. rally.
\#\#\# Answer choices: "}

With such clear instructions for error correction, it would be easier for LLM to perform context reasoning and choose the right answer for each blank than to predict the entire sentence from redundant N-best inputs.
In addition, the strong speech understanding ability of SpeechGPT enables ClozeGER to refer to source speech to make better choices.

\subsubsection{Logits Calibration with Prior Estimate}
\label{sssec:calibration}
Despite the promising performance, most recent works find that LLMs-based cloze test is vulnerable to option position changes due to their inherent ``selection bias''~\citep{kumar2022answer,wang2023large,pezeshkpour2023large,he2023investigating}.
Similarly, in our experiments, we have observed a strong ``selection bias'' towards option `A', especially in the cases where ClozeGER makes mistakes.
One reason is that most ground-truth options are `A' in the training data\footnote{Ablation study on debiased training is in Table~\ref{table:effect_calibration}.} as the 1-best hypothesis usually enjoys the best quality.
As a result, during inference when LLM find it hard to decide the answer choice, it tends to select `A' that can at least guarantee no performance drop, \emph{i.e.}, option `A' comes from 1-best hypothesis (baseline).
Inspired by prior works on permutation-based debiasing~\citep{wang2023large,zheng2023judging,zheng2023large}, we propose a logits calibration approach with prior estimation to alleviate this bias during inference.

Formally, we denote the question context as $c$, the $n$ option IDs (\emph{e.g.}, A/B/C) for one blank as $d$, and the default option contents (\emph{i.e.}, follow the order of N-best hypotheses) as $x$.
Take the second blank in Fig.~\ref{fig2} (c) as an example, we have $c =$ ``\textit{\textcolor[HTML]{666666}{[Blank1]}} he \textit{\textcolor[HTML]{666666}{[Blank2]}} need it'', $d =$ [A, B, C], $n = 3$, $x =$ [rarely, really, rally].
The concatenation of default option IDs and contents is denoted as $o$.

Then, we use $I$ to denote a permutation of $\{1, 2, \cdots, n\}$, and $\mathcal{I}$ to denote the set of all possible $I$s.
For better formulation, we denote $o^I$ as the concatenation of option IDs and $I$-permuted option contents, and $r_I(i)$ denotes the position of ID $i$ in $I$.
Take the above example to illustrate, assume $I = [2, 3, 1]$, then $o^I =$ \{A: really, B : rally, C : rarely\} and $r_I(1) = 3, r_I(2) = 1, r_I(3) = 2$.
In order to alleviate the selection bias of LLMs towards the option IDs, we have to first formulate it mathematically.
One feasible solution~\citep{wang2023large,zheng2023judging} is to enumerate all permutations of the option contents and average their output distributions for debiasing:
\begin{equation}
\label{eq4}
\begin{aligned}
    \mathcal{P}_\text{real} (x_i | c, o) &= \frac{1}{|\mathcal{I}|} \sum_{I\in\mathcal{I}} \mathcal{P}_\text{llm} (d_{r_I(i)} | c, o^I),
\end{aligned}
\end{equation}
where $i \in \{1, 2, \cdots, n\}$, and $\mathcal{P}_\text{real} (x_i|c, o)$ denotes the debiased (\emph{i.e.}, real) probability of $i$-th option content in $x$.
After such enumeration of all possible permutations of option contents, the ``selection bias'' towards option IDs could be well resolved.

Furthermore, considering calculating full permutations is prohibitively expensive ($\times n!$ costs), we leverage the cyclic permutation as an alternative, \emph{i.e.}, $I = \{(i, i+1, \cdots, n, 1, 2, \cdots, i-1)\}_{i=1}^n$.
Take the previous example, we have $I = \{(1, 2, 3), (2, 3, 1), (3, 1, 2)\}$.
It reduces the computational cost from $\times n!$ to $\times n$ and guarantees one pairing between each option ID and content.
However, the inference cost of $\times n$ is still much too high especially in practical scenarios.

Inspired by recent work on MCQ debiasing~\citep{zheng2023large}, it is a promising idea to disentangle the distribution bias of option IDs from the original predictions from LLMs.
The insight behind is that the option ID itself is inherently unrelated to the option contents, the option orders, and the context.
Therefore, the LLM predicted distribution over $d_i$ can be disentangled as a prior distribution of the option ID $d_i$ and the debiased (\emph{i.e.}, real) distribution of option content of $d_i$:
\begin{equation}
\label{eq5}
\begin{aligned}
    \mathcal{P}_\text{llm} (d_i | c, o) \propto \mathcal{P}_\text{prior} (d_i | c) \mathcal{P}_\text{real} (x_i | c, o),
\end{aligned}
\end{equation}
where we omit $I$ as only the default order of options needs to be considered during formal inference.
The prior distribution $\mathcal{P}_\text{prior} (d_i | c)$ indicates the LLM's selection bias towards option ID $d_i$, and the debiased distribution $\mathcal{P}_\text{real} (x_i | c, o)$ indicates the LLM's real confidence of option content $x_i$.

Inspired by recent work~\citep{zheng2023large}, we calculate the averaged prior distribution $\hat{\mathcal{P}}_\text{prior} (d_i)$ on validation set $\mathcal{D}_v$ to estimate $\mathcal{P}_\text{prior} (d_i | c)$.
In particular, we perform cyclic permutation $\mathcal{I}_c$ for each sample in $\mathcal{D}_v$ and send all of them for inference, and then we average their output distributions to obtain the prior distribution of option ID $d_i$:
\begin{equation}
\label{eq6}
\begin{aligned}
    \hat{\mathcal{P}}_\text{prior} (d_i) &= \frac{1}{|\mathcal{D}_v|}\sum_{\{c,o\}\in\mathcal{D}_v} \mathcal{P}_\text{prior} (d_i | c),\\
    \mathcal{P}_\text{prior} (d_i | c) &= \text{sm}\left(\frac{1}{|\mathcal{I}|} \sum_{I\in\mathcal{I}} \log\mathcal{P}_\text{llm} (d_i | c, o^I)\right),
\end{aligned}
\end{equation}
where ``sm'' denotes softmax operation.
With the estimated prior distribution, we can perform logits calibration during the inference stage:
\begin{equation}
\label{eq7}
\begin{aligned}
    \hat{\mathcal{P}}_\text{real} (x_i | c, o) \propto \mathcal{P}_\text{llm} (d_i | c, o) / \hat{\mathcal{P}}_{\text{prior}} (d_i),
\end{aligned}
\end{equation}
In case of the small size of $\mathcal{D}_v$, this logits calibration method would be efficient during inference.

\subsubsection{Post-processing}
\label{sssec:post_process}
After cloze test with logits calibration, many ASR errors captured by the blanks (\emph{i.e.}, varying tokens between N-best hypotheses) are corrected, but what about those remaining in the question context?
For example, as shown in Fig.~\ref{fig2} (c), the ASR model fails to recognize the word ``needs'', where all N-best hypotheses produce ``need''.
In this case, we need a simple post-processing stage to further correct them, with the following instructions:

\emph{"Below is a speech and its candidate transcription from a speech recognition system. 
Please listen to the speech and correct the candidate transcription.
\#\#\# Speech:\{speech units\}.
\#\#\# Candidate:\{cloze result\}.
\#\#\# Response: "}

Similar to GER, here we also use SpeechGPT with LoRA finetuning for post-processing.
This stage is necessary especially when ASR model does not perform well on current speech domains.

\section{Experiments}
\label{sec:exp}

\subsection{Setup}
\label{ssec:setup}

\noindent\textbf{Dataset.}
We utilize the HyPoradise (HP) dataset from the original GER benchmark~\citep{chen2023hp} for our experiments, which contains over 332K hypotheses-transcription pairs collected from multiple mainstream ASR corpora. 
Specifically, each transcription is paired with 5-best hypotheses transcribed from Whisper-Large model~\citep{radford2023robust} with beam search decoding. 
In this work, we select 9 popular ASR corpora from HyPoradise to evaluate the proposed ClozeGER, including WSJ~\citep{paul1992design}, CommonVoice~\citep{ardila2019common}, TED-LIUM3~\citep{hernandez2018ted}, SwitchBoard~\citep{godfrey1992switchboard}, LibriSpeech~\citep{panayotov2015librispeech}, CHiME-4~\citep{vincent2016chime4}, LRS2~\citep{chung2017lip}, ATIS~\citep{hemphill1990atis}, and CORAAL~\citep{kendall2021corpus}.
Since HyPoradise provides 5-best hypotheses for each sample, we follow it to set 5 options for each cloze blank.
More statistical details are in Appendix~\ref{asec:hp_details}.

\vspace{0.1cm}
\noindent\textbf{Models.}
As introduced before, we use SpeechGPT as the LLM in our main experiments, and later on we also try LLaMA-2-7b\footnote{\url{https://huggingface.co/meta-llama/Llama-2-7b-hf}}~\citep{touvron2023llama2} to verify the effectiveness of ClozeGER paradigm in case of no source speech input.
For efficient LLM finetuning, we employ the popular low-rank adapter (LoRA) tuning strategy~\citep{hu2021lora}, where the rank $r$ is set to 8 and the LoRA is added in the query, key, value, and output layers in each Transformer block~\citep{vaswani2017attention}. 
As a result, the number of trainable parameters is only 8.39 M, accounting for only 0.12\% of the LLM. 

\vspace{0.1cm}
\noindent\textbf{Training and Inference.}
During finetuning, we employ Adam optimizer~\citep{kingma2014adam} with a learning rate set to $2e^{-4}$ and warmup steps set to 100.
The number of training epochs is set to 5, the batch size is set to 256.
The maximum input sequence length is set to 1024.
For inference, we adopt top-$k$ and top-$p$ sampling strategies at the same time, where $k=40$ and $p=0.75$.
The temperature is set to 0.1, and beam size is set to 4.

\subsection{Main Results}
\label{ssec:main_results}

\begin{table*}[t]
\centering
\resizebox{1.0\textwidth}{!}{
\begin{tabular}{l|c|c|c|ccc|cc}
\toprule[1.2pt]
\multirow{2}{*}{Test Set} & \multirow{2}{*}{Baseline} & \emph{w/o Source Speech} & \multicolumn{4}{c|}{\emph{w/ Source Speech}} & \multicolumn{2}{c}{Oracle} \\
& & GER~\citeyearpar{chen2023hp} & GER & ClozeGER (ours) & + Calibration & + Post-processing & $o_{nb}$ & $o_{cp}$ \\
\midrule
WSJ & $4.2$ & $2.9_{\textcolor{teal}{-31.0\%}}$ & $2.7_{\textcolor{teal}{-35.7\%}}$ & $3.8_{\textcolor{teal}{-9.5\%}}$ & $3.3_{\textcolor{teal}{-21.4\%}}$ & $\bm{2.4_{\textcolor{teal}{-42.9\%}}}$ & $2.7$ & $1.0$ \\
CommonVoice & $14.4$ & $11.4_{\textcolor{teal}{-20.8\%}}$ & $10.1_{\textcolor{teal}{-29.9\%}}$ & $13.7_{\textcolor{teal}{-4.9\%}}$ & $12.4_{\textcolor{teal}{-13.9\%}}$ & $\bm{8.5_{\textcolor{teal}{-41.0\%}}}$ & $10.5$ & $7.5$ \\
TED-LIUM3 & $6.8$ & $5.8_{\textcolor{teal}{-14.7\%}}$ & $5.4_{\textcolor{teal}{-20.6\%}}$ & $6.1_{\textcolor{teal}{-10.3\%}}$ & $5.1_{\textcolor{teal}{-25.0\%}}$ & $\bm{4.8_{\textcolor{teal}{-29.4\%}}}$ & $4.4$ & $1.6$ \\
SwitchBoard & $16.4$ & $14.8_{\textcolor{teal}{-9.8\%}}$ & $14.3_{\textcolor{teal}{-12.8\%}}$ & $15.8_{\textcolor{teal}{-3.7\%}}$ & $15.0_{\textcolor{teal}{-8.5\%}}$ & $\bm{13.3_{\textcolor{teal}{-18.9\%}}}$ & $13.3$ & $4.6$ \\
LibriSpeech & $2.7$ & $2.7_{\textcolor{gray}{-0.0\%}}$ & $2.6_{\textcolor{teal}{-3.7\%}}$ & $2.7_{\textcolor{gray}{-0.0\%}}$ & $2.5_{\textcolor{teal}{-7.4\%}}$ & $\bm{2.5_{\textcolor{teal}{-7.4\%}}}$ & $1.9$ & $1.1$ \\
CHiME-4 & $9.4$ & $7.4_{\textcolor{teal}{-21.3\%}}$ & $7.9_{\textcolor{teal}{-16.0\%}}$ & $8.7_{\textcolor{teal}{-7.4\%}}$ & $7.6_{\textcolor{teal}{-19.1\%}}$ & $\bm{7.1_{\textcolor{teal}{-24.5\%}}}$ & $5.9$ & $2.7$ \\
LRS2 & $12.3$ & $10.7_{\textcolor{teal}{-13.0\%}}$ & $9.5_{\textcolor{teal}{-22.8\%}}$ & $10.7_{\textcolor{teal}{-13.0\%}}$ & $9.3_{\textcolor{teal}{-24.3\%}}$ & $\bm{7.6_{\textcolor{teal}{-38.2\%}}}$ & $7.5$ & $2.8$ \\
ATIS & $7.3$ & $2.9_{\textcolor{teal}{-60.3\%}}$ & $2.4_{\textcolor{teal}{-67.1\%}}$ & $7.1_{\textcolor{teal}{-2.7\%}}$ & $6.5_{\textcolor{teal}{-11.0\%}}$ & $\bm{2.1_{\textcolor{teal}{-71.2\%}}}$ & $4.1$ & $1.0$ \\
CORAAL & $29.2$ & $27.9_{\textcolor{teal}{-4.5\%}}$ & $27.6_{\textcolor{teal}{-5.5\%}}$ & $29.1_{\textcolor{teal}{-0.3\%}}$ & $28.1_{\textcolor{teal}{-3.8\%}}$ & $\bm{26.7_{\textcolor{teal}{-8.6\%}}}$ & $27.9$ & $10.9$ \\
\bottomrule[1.2pt]
\end{tabular}}
\vspace{-0.1cm}
\caption{WER (\%) results of ClozeGER with SpeechGPT and LoRA. 
``+ Calibration'' denotes adding logits calibration on ClozeGER to remove the selection bias, and ``+ Post-processing'' denotes further adding the post-processing stage to correct the context.
$o_{nb}$ denotes the N-best oracle that refers to word error rate (WER) of the ``best candidate'' in the N-best list, and $o_{cp}$ denotes the compositional oracle that is the best achievable WER using all the tokens in N-best hypotheses.
They indicate the upper-bounds of LM rescoring and GER (with occurred tokens), respectively.
The subscript percentage denotes the relative WER reduction over ASR baseline.
}
\vspace{-0.1cm}
\label{table:main_results}
\end{table*}

\begin{table*}[t]
\vspace{-0.1cm}
\centering
\resizebox{1.0\textwidth}{!}{
\begin{tabular}{p{2.8cm}|c|c|ccc|cc}
\toprule[1.2pt]
\multirow{2}{*}{Test Set} & \multirow{2}{*}{Baseline} & \multicolumn{4}{c|}{\emph{w/o Source Speech}} & \multicolumn{2}{c}{Oracle} \\
& & GER~\citeyearpar{chen2023hp} & ClozeGER (ours) & + Calibration & + Post-processing & $o_{nb}$ & $o_{cp}$ \\
\midrule
WSJ & $4.2$ & $2.8_{\textcolor{teal}{-33.3\%}}$ & $3.7_{\textcolor{teal}{-11.9\%}}$ & $3.3_{\textcolor{teal}{-21.4\%}}$ & $\bm{2.5_{\textcolor{teal}{-40.5\%}}}$ & $2.7$ & $1.0$ \\
CommonVoice & $14.4$ & $10.8_{\textcolor{teal}{-25.0\%}}$ & $13.1_{\textcolor{teal}{-9.0\%}}$ & $12.4_{\textcolor{teal}{-13.9\%}}$ & $\bm{8.6_{\textcolor{teal}{-40.3\%}}}$ & $10.5$ & $7.5$ \\
TED-LIUM3 & $6.8$ & $5.3_{\textcolor{teal}{-22.1\%}}$ & $6.0_{\textcolor{teal}{-11.8\%}}$ & $5.0_{\textcolor{teal}{-26.5\%}}$ & $\bm{4.7_{\textcolor{teal}{-30.9\%}}}$ & $4.4$ & $1.6$ \\
SwitchBoard & $16.4$ & $14.6_{\textcolor{teal}{-11.0\%}}$ & $15.6_{\textcolor{teal}{-4.9\%}}$ & $14.5_{\textcolor{teal}{-11.6\%}}$ & $\bm{12.9_{\textcolor{teal}{-21.3\%}}}$ & $13.3$ & $4.6$ \\
LibriSpeech & $2.7$ & $2.7_{\textcolor{gray}{-0.0\%}}$ & $2.6_{\textcolor{teal}{-3.7\%}}$ & $2.4_{\textcolor{teal}{-11.1\%}}$ & $\bm{2.4_{\textcolor{teal}{-11.1\%}}}$ & $1.9$ & $1.1$ \\
CHiME-4 & $9.4$ & $7.3_{\textcolor{teal}{-22.3\%}}$ & $7.9_{\textcolor{teal}{-16.0\%}}$ & $7.2_{\textcolor{teal}{-23.4\%}}$ & $\bm{7.0_{\textcolor{teal}{-25.5\%}}}$ & $5.9$ & $2.7$ \\
LRS2 & $12.3$ & $10.5_{\textcolor{teal}{-14.6\%}}$ & $10.5_{\textcolor{teal}{-14.6\%}}$ & $9.0_{\textcolor{teal}{-26.8\%}}$ & $\bm{7.4_{\textcolor{teal}{-39.8\%}}}$ & $7.5$ & $2.8$ \\
ATIS & $7.3$ & $2.4_{\textcolor{teal}{-67.1\%}}$ & $6.3_{\textcolor{teal}{-13.7\%}}$ & $5.8_{\textcolor{teal}{-20.5\%}}$ & $\bm{2.1_{\textcolor{teal}{-71.2\%}}}$ & $4.1$ & $1.0$ \\
CORAAL & $29.2$ & $27.4_{\textcolor{teal}{-6.2\%}}$ & $29.1_{\textcolor{teal}{-0.3\%}}$ & $27.9_{\textcolor{teal}{-4.5\%}}$ & $\bm{26.8_{\textcolor{teal}{-8.2\%}}}$ & $27.9$ & $10.9$ \\
\bottomrule[1.2pt]
\end{tabular}}
\vspace{-0.1cm}
\caption{WER (\%) results of ClozeGER with LLaMA-2-7b and LoRA.
This study investigates the performance of our ClozeGER in case of no source speech input.
$o_{nb}$ and $o_{cp}$ follow the same definitions as those in Table~\ref{table:main_results}.
}
\label{table:main_results_no_speech}
\vspace{-0.3cm}
\end{table*}

Table~\ref{table:main_results} presents the WER results of ClozeGER with SpeechGPT and LoRA tuning.
First, we can observe that vanilla GER achieves significant improvements over Whisper ASR baseline, and introducing source speech as extra input further enhances the performance.
In comparison, our proposed ClozeGER also improves the baseline but underperforms the GER approach.
There are two reasons, the cloze test suffers from selection bias and cannot yield satisfactory results, and on the other hand, there are many errors exist in the cloze context due to imperfect N-best list quality (\emph{i.e.}, Whisper is a general ASR model and may not perform well in every specific domain).
To this end, we first propose a logits calibration approach to alleviate the selection bias, which results in considerable WER reductions.
Furthermore, we add a post-processing stage to correct the errors in cloze context, which moves one step forward and outperforms the GER approach with source speech input, where some results even surpass the N-best oracle.


Table~\ref{table:main_results_no_speech} reports the WER results of ClozeGER using LLaMA-2 as a backbone in case of no source speech as input, where we observe similar gains of performance of the proposed ClozeGER over GER baseline.
It demonstrates the general effectiveness of ClozeGER paradigm, as well as the proposed logits calibration and post-processing techniques.


\begin{table*}[t]
\vspace{-0.1cm}
\centering
\resizebox{1.0\textwidth}{!}{
\begin{tabular}{p{3.6cm}|c|c|c|c|c}
\toprule[1.2pt]
Label Dist. / Prior (\%) & A & B & C & D & E \\
\midrule
WSJ & \hspace{0.15cm}$75.51$\hspace{0.15cm} / \hspace{0.15cm}$95.68$\hspace{0.15cm} & \hspace{0.15cm}$10.27$\hspace{0.15cm} / \hspace{0.15cm}$2.92$\hspace{0.15cm} & \hspace{0.15cm}$6.84$\hspace{0.15cm} / \hspace{0.15cm}$0.63$\hspace{0.15cm} & \hspace{0.15cm}$4.16$\hspace{0.15cm} / \hspace{0.15cm}$0.42$\hspace{0.15cm} & \hspace{0.15cm}$3.22$\hspace{0.15cm} / \hspace{0.15cm}$0.35$\hspace{0.15cm} \\
CommonVoice & \hspace{0.15cm}$80.32$\hspace{0.15cm} / \hspace{0.15cm}$95.94$\hspace{0.15cm} & \hspace{0.15cm}$8.33$\hspace{0.15cm} / \hspace{0.15cm}$2.54$\hspace{0.15cm} & \hspace{0.15cm}$5.47$\hspace{0.15cm} / \hspace{0.15cm}$0.59$\hspace{0.15cm} & \hspace{0.15cm}$3.34$\hspace{0.15cm} / \hspace{0.15cm}$0.57$\hspace{0.15cm} & \hspace{0.15cm}$2.54$\hspace{0.15cm} / \hspace{0.15cm}$0.36$\hspace{0.15cm} \\
TED-LIUM3 & \hspace{0.15cm}$75.22$\hspace{0.15cm} / \hspace{0.15cm}$98.13$\hspace{0.15cm} & \hspace{0.15cm}$10.34$\hspace{0.15cm} / \hspace{0.15cm}$1.63$\hspace{0.15cm} & \hspace{0.15cm}$6.89$\hspace{0.15cm} / \hspace{0.15cm}$0.17$\hspace{0.15cm} & \hspace{0.15cm}$4.30$\hspace{0.15cm} / \hspace{0.15cm}$0.03$\hspace{0.15cm} & \hspace{0.15cm}$3.25$\hspace{0.15cm} / \hspace{0.15cm}$0.04$\hspace{0.15cm} \\
SwitchBoard & \hspace{0.15cm}$77.93$\hspace{0.15cm} / \hspace{0.15cm}$96.70$\hspace{0.15cm} & \hspace{0.15cm}$9.18$\hspace{0.15cm} / \hspace{0.15cm}$2.78$\hspace{0.15cm} & \hspace{0.15cm}$6.07$\hspace{0.15cm} / \hspace{0.15cm}$0.30$\hspace{0.15cm} & \hspace{0.15cm}$3.85$\hspace{0.15cm} / \hspace{0.15cm}$0.09$\hspace{0.15cm} & \hspace{0.15cm}$2.98$\hspace{0.15cm} / \hspace{0.15cm}$0.13$\hspace{0.15cm} \\
LibriSpeech & \hspace{0.15cm}$73.80$\hspace{0.15cm} / \hspace{0.15cm}$98.08$\hspace{0.15cm} & \hspace{0.15cm}$11.51$\hspace{0.15cm} / \hspace{0.15cm}$1.50$\hspace{0.15cm} & \hspace{0.15cm}$7.63$\hspace{0.15cm} / \hspace{0.15cm}$0.28$\hspace{0.15cm} & \hspace{0.15cm}$4.01$\hspace{0.15cm} / \hspace{0.15cm}$0.08$\hspace{0.15cm} & \hspace{0.15cm}$3.05$\hspace{0.15cm} / \hspace{0.15cm}$0.06$\hspace{0.15cm} \\
CHiME-4 & \hspace{0.15cm}$77.56$\hspace{0.15cm} / \hspace{0.15cm}$84.17$\hspace{0.15cm} & \hspace{0.15cm}$8.83$\hspace{0.15cm} / \hspace{0.15cm}$9.13$\hspace{0.15cm} & \hspace{0.15cm}$6.89$\hspace{0.15cm} / \hspace{0.15cm}$4.54$\hspace{0.15cm} & \hspace{0.15cm}$3.98$\hspace{0.15cm} / \hspace{0.15cm}$1.33$\hspace{0.15cm} & \hspace{0.15cm}$2.73$\hspace{0.15cm} / \hspace{0.15cm}$0.83$\hspace{0.15cm} \\
LRS2 & \hspace{0.15cm}$77.21$\hspace{0.15cm} / \hspace{0.15cm}$95.85$\hspace{0.15cm} & \hspace{0.15cm}$9.81$\hspace{0.15cm} / \hspace{0.15cm}$3.63$\hspace{0.15cm} & \hspace{0.15cm}$6.58$\hspace{0.15cm} / \hspace{0.15cm}$0.31$\hspace{0.15cm} & \hspace{0.15cm}$3.54$\hspace{0.15cm} / \hspace{0.15cm}$0.15$\hspace{0.15cm} & \hspace{0.15cm}$2.86$\hspace{0.15cm} / \hspace{0.15cm}$0.07$\hspace{0.15cm} \\
ATIS & \hspace{0.15cm}$78.43$\hspace{0.15cm} / \hspace{0.15cm}$82.78$\hspace{0.15cm} & \hspace{0.15cm}$10.08$\hspace{0.15cm} / \hspace{0.15cm}$8.67$\hspace{0.15cm} & \hspace{0.15cm}$5.63$\hspace{0.15cm} / \hspace{0.15cm}$4.27$\hspace{0.15cm} & \hspace{0.15cm}$3.08$\hspace{0.15cm} / \hspace{0.15cm}$2.28$\hspace{0.15cm} & \hspace{0.15cm}$2.78$\hspace{0.15cm} / \hspace{0.15cm}$2.00$\hspace{0.15cm} \\
CORAAL & \hspace{0.15cm}$77.58$\hspace{0.15cm} / \hspace{0.15cm}$83.08$\hspace{0.15cm} & \hspace{0.15cm}$8.85$\hspace{0.15cm} / \hspace{0.15cm}$7.86$\hspace{0.15cm} & \hspace{0.15cm}$6.21$\hspace{0.15cm} / \hspace{0.15cm}$3.78$\hspace{0.15cm} & \hspace{0.15cm}$3.95$\hspace{0.15cm} / \hspace{0.15cm}$2.73$\hspace{0.15cm} & \hspace{0.15cm}$3.40$\hspace{0.15cm} / \hspace{0.15cm}$2.54$\hspace{0.15cm} \\
\bottomrule[1.2pt]
\end{tabular}}
\caption{Training label distribution (\%) and the estimated prior distribution (\%) over 5 option IDs (\emph{i.e.}, `A', `B', `C', `D', and `E') of different datasets.
The training label distribution refers to the proportions of each option ID in the labels of cloze training data, and the estimated prior distribution is illustrated in Eq.(\ref{eq6}) as $\hat{\mathcal{P}}_\text{prior} (d_i)$.
}
\label{table:prior_dist}
\vspace{-0.2cm}
\end{table*}

\begin{figure}[t]
\begin{center}
\includegraphics[width=0.973\columnwidth]{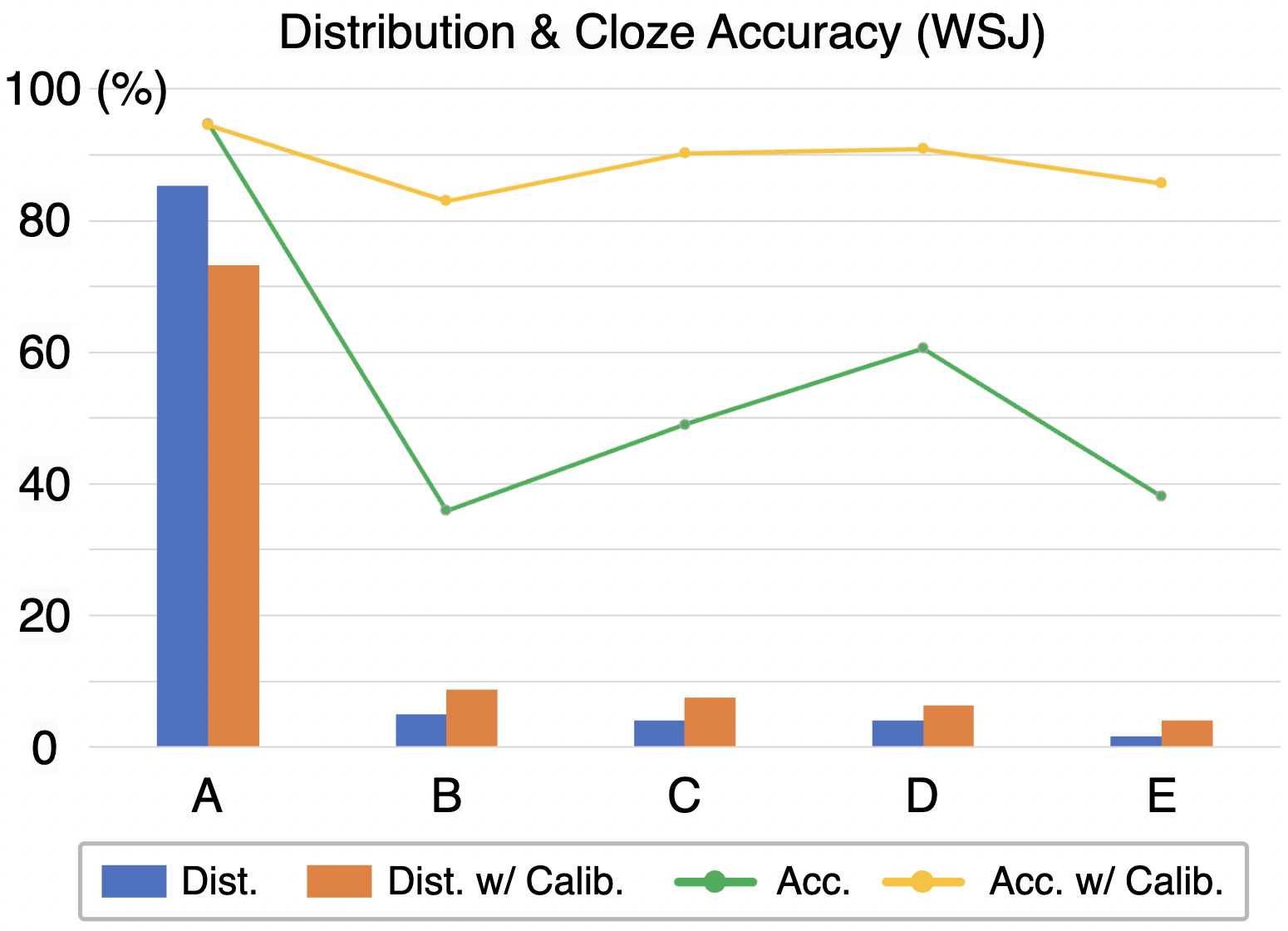}
\end{center}
\vspace{-0.2cm}
\caption{Distribution and cloze accuracy of five options with logits calibration on WSJ dataset.
``Dist.'' denotes the distribution of five options in the predictions, ``Acc.'' denotes the predicting accuracy of each ground-truth option, and ``w/ Calib.'' denotes with logits calibration.
}
\label{fig3}
\vspace{-0.2cm}
\end{figure}

\begin{table}[t]
\centering
\resizebox{1.0\columnwidth}{!}{
\begin{tabular}{l|cc|cc}
\toprule[1.2pt]
\multirow{2}{*}{Test Set} & \multicolumn{2}{c|}{ClozeGER} & \multicolumn{2}{c}{ClozeGER \emph{w/ Calib.}} \\
& Acc (\%) & WER (\%) & Acc (\%) & WER(\%) \\
\midrule
WSJ & $84.7$ & $3.8$ & $92.6$ & $3.3$ \\
CommonVoice & $82.6$ & $13.7$ & $91.5$ & $12.4$ \\
TED-LIUM3 & $76.0$ & $6.1$ & $91.3$ & $5.1$ \\
SwitchBoard & $78.7$ & $15.8$ & $86.8$ & $15.0$ \\
LibriSpeech & $80.3$ & $2.7$ & $95.3$ & $2.5$ \\
CHiME-4 & $74.4$ & $8.7$ & $87.5$ & $7.6$ \\
LRS2 & $83.9$ & $10.7$ & $91.7$ & $9.3$ \\
ATIS & $78.9$ & $7.1$ & $84.5$ & $6.5$ \\
CORAAL & $78.9$ & $29.1$ & $87.3$ & $28.1$ \\
\bottomrule[1.2pt]
\end{tabular}}
\caption{Effect of the logits calibration approach in proposed ClozeGER framework, in terms of the cloze test accuracy and final WER performance.
}
\label{table:effect_calibration}
\vspace{-0.3cm}
\end{table}

\subsection{Ablation Study and Analysis}
\label{ssec:ablation_analysis}

\noindent\textbf{Why we need logits calibration?} \par
\noindent We note that ClozeGER only produces limited improvement and even underperforms the GER baseline, where one key reason is the ``selection bias'' towards option IDs.
Take the WSJ dataset as an example, we visualize the distribution of predicted option IDs in Fig.~\ref{fig3}, where over 80\% of predictions fall on option `A'.
This phenomenon can be explained by the imbalanced training label distribution\footnote{Because top-1 hypothesis enjoys the best quality and is most likely to be the ground-truth choice.} and its resulted ``selection bias'' as shown in Table~\ref{table:prior_dist}.
As a result, the proposed ClozeGER yields poor predicting accuracy on options `B' to `E', which limits its final WER performance.

\vspace{0.1cm}
\noindent\textbf{How does logits calibration work?} \par
\noindent To alleviate this limitation, we propose to estimate a prior distribution to represent the selection bias and then remove it from the output logits during inference stage to conduct calibration.
As illustrated by the orange bars and yellow lines in Fig.~\ref{fig3}, our proposed calibration approach mitigates the imbalance of predicted options and effectively improves their cloze accuracy.
As a result, the overall cloze accuracy is significantly improved on various ASR datasets in Table~\ref{table:effect_calibration}, which thus produces better WER performance in the final.
More visualization on other datasets is in the Appendix Fig.~\ref{fig5}.

\vspace{0.1cm}
\noindent\textbf{Why not do calibration during training stage?} \par
\noindent One may raise concerns about why we conduct the calibration during inference instead of training stage, where simply shuffling the options seems able to mitigate the selection bias.
To this end, we present an ablation study in Table~\ref{table:cali_train_infer} to explore calibration in different stages, where we observe that shuffled training can indeed achieve some improvement over ClozeGER but still lag behind the proposed logits calibration approach.
When further adding the post-processing stage, our calibration also produces better performance, indicating that it is more beneficial to remove selection bias during the inference stage rather than training stage.

This observation may suggest that within our specific framework, the model's acquisition of selection bias during training is \textit{somehow advantageous}, as the strategy of arbitrarily selecting `A' would at worst regress to the baseline (top-1 hypothesis) without deterioration.
As a result, the task difficulty of ClozeGER is naturally reduced, because it can rely on the heuristic of selecting `A' when feeling uncertain and hard to make the choice.
Thereafter, the logits calibration during inference removes such bias by diversifying some of the ‘A’ choices to other options to improve the accuracy.

\begin{table*}[t]
\centering
\resizebox{1.0\textwidth}{!}{
\begin{tabular}{p{2.8cm}|c|c|c|cc|cc|cc}
\toprule[1.2pt]
\multirow{2}{*}{Test Set} & \multirow{2}{*}{Baseline} & \multirow{2}{*}{\hspace{0.1cm}GER\hspace{0.1cm}} & \multirow{2}{*}{ClozeGER} & \multicolumn{2}{c|}{Train stage} & \multicolumn{2}{c|}{Infer stage} & \multicolumn{2}{c}{Oracle} \\
& & & & + Shuf. & + Post. & + Calib. & + Post. & $o_{nb}$ & $o_{cp}$ \\
\midrule
WSJ & $4.2$ & $2.7$ & $3.8$ & $3.5$ & $2.7$ & $3.3$ & $\bm{2.4_{\textcolor{teal}{-42.9\%}}}$ & $2.7$ & $1.0$ \\
CommonVoice & $14.4$ & $10.1$ & $13.7$ & $12.9$ & $9.2$ & $12.4$ & $\bm{8.5_{\textcolor{teal}{-41.0\%}}}$ & $10.5$ & $7.5$ \\
TED-LIUM3 & $6.8$ & $5.4$ & $6.1$ & $5.7$ & $5.1$ & $5.1$ & $\bm{4.8_{\textcolor{teal}{-29.4\%}}}$ & $4.4$ & $1.6$ \\
SwitchBoard & $16.4$ & $14.3$ & $15.8$ & $15.3$ & $13.8$ & $15.0$ & $\bm{13.3_{\textcolor{teal}{-18.9\%}}}$ & $13.3$ & $4.6$ \\
LibriSpeech & $2.7$ & $2.6$ & $2.7$ & $2.6$ & $2.5$ & $2.5$ & $\bm{2.5_{\textcolor{teal}{-7.4\%}}}$ & $1.9$ & $1.1$ \\
CHiME-4 & $9.4$ & $7.9$ & $8.7$ & $8.0$ & $7.3$ & $7.6$ & $\bm{7.1_{\textcolor{teal}{-24.5\%}}}$ & $5.9$ & $2.7$ \\
LRS2 & $12.3$ & $9.5$ & $10.7$ & $9.8$ & $7.9$ & $9.3$ & $\bm{7.6_{\textcolor{teal}{-38.2\%}}}$ & $7.5$ & $2.8$ \\
ATIS & $7.3$ & $2.4$ & $7.1$ & $6.9$ & $2.4$ & $6.5$ & $\bm{2.1_{\textcolor{teal}{-71.2\%}}}$ & $4.1$ & $1.0$ \\
CORAAL & $29.2$ & $27.6$ & $29.1$ & $28.4$ & $27.2$ & $28.1$ & $\bm{26.7_{\textcolor{teal}{-8.6\%}}}$ & $27.9$ & $10.9$ \\
\bottomrule[1.2pt]
\end{tabular}}
\caption{Effect of calibration during different stages, \emph{i.e.}, training and inference.
``+ Shuf.'' denotes shuffling the option contents during training stage (keep the order of option IDs as ``A, B, C, D, E''), ``+ Calib.'' denotes using logits calibration during inference stage, ``+ Post.'' denotes adding pre-processing on top of shuffling or calibration.
}
\label{table:cali_train_infer}
\end{table*}

\begin{table*}[t]
\centering
\resizebox{1.0\textwidth}{!}{
\begin{tabular}{p{2.8cm}|c|cc|cc|cc}
\toprule[1.2pt]
\multirow{2}{*}{Test Set} & \multirow{2}{*}{Baseline} & \multicolumn{2}{c|}{GER} & \multicolumn{2}{c|}{ClozeGER \emph{w/ Calib.}} & \multicolumn{2}{c}{Oracle} \\
& & Original & + Post-processing & Original & + Post-processing & $o_{nb}$ & $o_{cp}$ \\
\midrule
WSJ & $4.2$ & $2.7_{\textcolor{teal}{-35.7\%}}$ & $2.6_{\textcolor{teal}{-38.1\%}}$ & $3.3_{\textcolor{teal}{-21.4\%}}$ & $\bm{2.4_{\textcolor{teal}{-42.9\%}}}$ & $2.7$ & $1.0$ \\
CommonVoice & $14.4$ & $10.1_{\textcolor{teal}{-29.9\%}}$ & $9.6_{\textcolor{teal}{-33.3\%}}$ & $12.4_{\textcolor{teal}{-13.9\%}}$ & $\bm{8.5_{\textcolor{teal}{-41.0\%}}}$ & $10.5$ & $7.5$ \\
TED-LIUM3 & $6.8$ & $5.4_{\textcolor{teal}{-20.6\%}}$ & $5.2_{\textcolor{teal}{-23.5\%}}$ & $5.1_{\textcolor{teal}{-25.0\%}}$ & $\bm{4.8_{\textcolor{teal}{-29.4\%}}}$ & $4.4$ & $1.6$ \\
SwitchBoard & $16.4$ & $14.3_{\textcolor{teal}{-12.8\%}}$ & $14.0_{\textcolor{teal}{-14.6\%}}$ & $15.0_{\textcolor{teal}{-8.5\%}}$ & $\bm{13.3_{\textcolor{teal}{-18.9\%}}}$ & $13.3$ & $4.6$ \\
LibriSpeech & $2.7$ & $2.6_{\textcolor{teal}{-3.7\%}}$ & $2.6_{\textcolor{teal}{-3.7\%}}$ & $2.5_{\textcolor{teal}{-7.4\%}}$ & $\bm{2.5_{\textcolor{teal}{-7.4\%}}}$ & $1.9$ & $1.1$ \\
CHiME-4 & $9.4$ & $7.9_{\textcolor{teal}{-16.0\%}}$ & $7.8_{\textcolor{teal}{-17.0\%}}$ & $7.6_{\textcolor{teal}{-19.1\%}}$ & $\bm{7.1_{\textcolor{teal}{-24.5\%}}}$ & $5.9$ & $2.7$ \\
LRS2 & $12.3$ & $9.5_{\textcolor{teal}{-22.8\%}}$ & $9.0_{\textcolor{teal}{-26.8\%}}$ & $9.3_{\textcolor{teal}{-24.3\%}}$ & $\bm{7.6_{\textcolor{teal}{-38.2\%}}}$ & $7.5$ & $2.8$ \\
ATIS & $7.3$ & $2.4_{\textcolor{teal}{-67.1\%}}$ & $2.3_{\textcolor{teal}{-68.5\%}}$ & $6.5_{\textcolor{teal}{-11.0\%}}$ & $\bm{2.1_{\textcolor{teal}{-71.2\%}}}$ & $4.1$ & $1.0$ \\
CORAAL & $29.2$ & $27.6_{\textcolor{teal}{-5.5\%}}$ & $27.3_{\textcolor{teal}{-6.5\%}}$ & $28.1_{\textcolor{teal}{-3.8\%}}$ & $\bm{26.7_{\textcolor{teal}{-8.6\%}}}$ & $27.9$ & $10.9$ \\
\bottomrule[1.2pt]
\end{tabular}}
\caption{Effect of the post-processing stage on GER and our proposed ClozeGER frameworks (with SpeechGPT as LLM).
``Calib.'' denotes the logits calibration approach.
$o_{nb}$ and $o_{cp}$ follow the same definitions as those in Table~\ref{table:main_results}.
}
\label{table:effect_postprocess}
\end{table*}

\vspace{0.1cm}
\noindent\textbf{Why we need post-processing?} \par
\noindent Table~\ref{table:effect_postprocess} further investigates the role of the post-processing stage in ClozeGER paradigm.
We observe that such post-processing is necessary to further correct the errors in cloze context, which results in promising gains of performance.
On the other hand, this phenomenon also reflects the sub-optimal quality of N-best hypotheses, according to the errors in cloze context (see Fig.~\ref{fig2} (c)), as Whisper is a general ASR model that may not generalize well to some specific domains like accents.

\vspace{0.1cm}
\noindent\textbf{The role of ClozeGER and post-processing.} \par
\noindent One may raise concerns on whether the effectiveness of our approach is all attributed to post-processing.
To this end, we add it onto the GER baseline, which also shows some improvement but still underperforms our ClozeGER, indicating that our proposed ClozeGER paradigm and logits calibration raises the upper-bound performance of GER by correcting errors in a targeted manner.
Based on that, the post-processing aims to further correct the errors in cloze context that cannot be resolved by the cloze-test paradigm.

\vspace{0.1cm}
\noindent\textbf{Case study.} \par
\noindent We illustrate a case study in Fig.~\ref{fig2} to interpret the motivation of our approach.
First, we introduce source speech as extra input to improve output fidelity, \emph{i.e.}, avoid removing the word ``Think''.
Second, we reformat GER as a cloze test to reduce the task difficulty with clear instructions, \emph{i.e.}, explicitly prompt the LLM to select a word from [``rarely'', ``really'', ``rally''], which results in an effective correction.
Finally, we note that there still exist some errors in the cloze context, \emph{e.g.}, ``need'', which cannot be corrected by the cloze-test paradigm.
To this end, we design a post-processing stage to further remove them and improve the final output.

\section{Conclusion}
\label{sec:conclusion}
In this paper, we propose ClozeGER, a new paradigm for ASR generative error correction.
First, we introduce a multimodal LLM (\emph{i.e.}, SpeechGPT) to receive source speech as extra input to improve the fidelity of correction output.
Then, we reformat GER as a cloze test with logits calibration to remove the input information redundancy and simplify GER with clear instructions.
Experimental evidence shows that ClozeGER achieves a new breakthrough over vanilla GER on 9 popular ASR datasets.
Further analysis verifies the effectiveness of different modules in our framework.

\section*{Limitations}
This work introduces an extra input of source speech to improve the output fidelity, which achieves some improvements but is somewhat limited since we only employ a new prompt for LLMs to exploit the source speech.
In future, we may investigate more advanced multimodal prompting techniques for it and also combine them with our proposed cloze-test paradigm for further improvements.
In addition, we believe it should also be beneficial to further investigate the reasons for the sub-optimal performance of cloze-test paradigm, as well as integrate the calibration and post-processing stages as an end-to-end pipeline in future work.

\section*{Ethics Statement}
This work does not pose any ethical issues.
All the data used in this paper are publicly available and under the following licenses: the Creative Commons BY-NC-ND 3.0 License, Creative Commons BY-NC-ND 4.0 License, Creative Commons BY-NC-SA 4.0 License, Creative Commons Attribution 4.0 International License, Creative Commons (CC0) License, the LDC User Agreement for Non-Members, the TED Terms of Use, the YouTube's Terms of Service, and the BBC’s Terms of Use.

\bibliography{anthology,custom}

\appendix

\begin{table*}[t]
\centering
\resizebox{1.0\textwidth}{!}{
\begin{tabular}{cc|ccc|ccc}
\toprule[1.2pt]
\multicolumn{2}{c|}{Domain} & \multirow{2}{*}{Training Set}& \multirow{2}{*}{\# Pairs}&\multirow{2}{*}{Length} & \multirow{2}{*}{Test Set} & \multirow{2}{*}{\# Pairs} &\multirow{2}{*}{Length}\\
Source & Category &  &  &  & & & \\
\midrule
WSJ & Business News & \emph{train-si284} & 37,514 & 17.5 & \emph{dev93 \& eval92} & 836 & 16.9 \\
CommonVoice & Speaker Accents & \emph{train-accent} & 49,758 & 10.5 & \emph{test-accent} & 2,000 & 10.5 \\
TED-LIUM3 & TED Talks & \emph{train} & 47,500 & 12.6 & \emph{test} & 2,500 & 12.6\\
SwitchBoard & Telephone & \emph{train} & 36,539 & 11.8 & \emph{eval2000} & 2,000 & 11.8\\
LibriSpeech & Audiobooks & \emph{train-960} & 88,200 & 33.7 & \emph{test-clean} & 2,620 & 20.1 \\
CHiME4 & Background Noise & \emph{tr05-real-noisy} & 9,600 & 17.0 & \emph{test-real} & 1,320 & 16.4 \\
LRS2 & BBC Television & \emph{train} & 42,940 & 7.6 & \emph{test} & 2,259 & 7.6 \\
ATIS & Airline Info. & \emph{train} & 3,964 & 12.4 & \emph{test} & 809  & 11.3 \\
CORAAL & Interview & \emph{train} & 1,728 & 24.2 & \emph{test} & 100 & 24.0  \\
\midrule
\multicolumn{2}{c|}{Total} & \emph{train} & 317,743 & 18.1 & \emph{test}& 14,444 & 13.4 \\
\bottomrule[1.2pt]
\end{tabular}}
\caption{HyPoradise dataset statistics in terms of different ASR domains (\emph{i.e.}, including speech and text domains), the number of hypotheses-transcription pairs, and the average utterance length of each dataset.
}
\label{table:statistics}
\end{table*}

\section{HyPoradise Dataset Details}
\label{asec:hp_details}

\subsection{Hypotheses Generation}
\label{assec:hypo_generation}
We employ the HyPoradise (HP) dataset\footnote{\url{https://huggingface.co/datasets/PeacefulData/HP-v0}} from original GER benchmark~\citep{chen2023hp}, which contains over 332K pairs of N-best hypotheses and ground-truth transcription.
The hypotheses are generated using Whisper-Large~\citep{radford2023robust} beam search decoding, where the beam size is set to 50.
After removing repetitive utterances, the top-5 hypotheses with the highest probabilities are selected as the final N-best list.
The HyPoradise dataset is built by carrying out this decoding strategy on multiple popular ASR datasets as introduced in~\S\ref{assec:asr_corpora_select}.
As a result, the detailed statistics of HyPoradise dataset is illustrated in Table~\ref{table:statistics}.

\subsection{ASR Corpora Selection}
\label{assec:asr_corpora_select}
For ASR corpora selection, we follow original benchmark~\cite{chen2023hp} to cover common ASR scenarios, \emph{e.g.}, noise and accents.
Consequently, the following corpora with evident domain characteristics are collected to build the HP dataset.

\vspace{0.1cm}
\noindent\textbf{WSJ}~\cite{paul1992design}: The Wall Street Journal (WSJ) is a widely-used benchmark for speech recognition. 
It includes read speech from speakers in a controlled environment, with a focus on business news and financial data.
The \emph{train-si284} split (37,514 samples) is utilized to generate HP training set.
The \emph{dev93} (503 samples) and \emph{eval92} (333 samples) splits are combined to build test set.

\vspace{0.1cm}
\noindent\textbf{CommonVoice}~\cite{ardila2019common}: 
CommonVoice 5.1 is a publicly available dataset for automatic speech recognition. 
It contains speech recordings from diverse speakers in over 60 languages.
To generate HP dataset, they randomly select 51,758 samples from its \emph{train-en} split with various accent labels, including African, Australian, Indian, and Singaporean.
Then, it is separated into two parts to build training (with 49,758 samples) and test (with 2,000 samples) sets respectively.

\vspace{0.1cm}
\noindent\textbf{TED-LIUM3}~\cite{hernandez2018ted}: TED-LIUM3 is a speech dataset recorded from TED talks. It contains a diverse range of background noise, speaker accents, and speech topics.
Considering its large size, they randomly select 50,000 samples from its \emph{train} split for HP dataset generation, which is then separated into training (47,500 samples) and test (2,500 samples) sets. 

\vspace{0.1cm}
\noindent\textbf{SwitchBoard}~\cite{godfrey1992switchboard}: The SwitchBoard corpus is a telephone speech dataset collected from conversations. 
It focuses on North American English and involves over 2,400 conversations from around 200 speakers.
They randomly select 36,539 samples from its \emph{train} split to generate HP training set, as well as 2,000 samples from the \emph{eval2000} split to generate HP test set.

\vspace{0.1cm}
\noindent\textbf{LibriSpeech}~\cite{panayotov2015librispeech}: LibriSpeech is a public corpus of read speech from audiobooks, including 1,000 hours of labeled speech data from diverse speakers, genders, and accents. 
To generate HP training data, they exclude some simple utterances from its \emph{train-960} split that yield 0\% WER, which results in 88,200 training samples.
The \emph{test-clean} split (2,620 samples) is used for HP test data.

\vspace{0.1cm}
\noindent\textbf{CHiME-4}~\cite{vincent2016chime4}: CHiME-4 is a dataset for far-field noisy speech recognition~\cite{chen2022noise,chen2022self,chen2023leveraging,chen2023unsupervised,hu2022interactive,hu2022dual,hu2023gradient,hu2023wav2code,hu2023hearing,hu2023mir,hu2023cross,zhu2023robust,zhu2024multichannel}. 
It includes real and simulated noisy recordings in four noisy environments, \textit{i.e.}, bus, cafe, pedestrian area, and street junction.
Its \emph{tr05-real-noisy} (9,600 samples) and \emph{test-real} (1,320 samples) splits are used to generate HP training and test data, respectively.

\begin{figure}[t]
\begin{center}
\includegraphics[width=1.0\columnwidth]{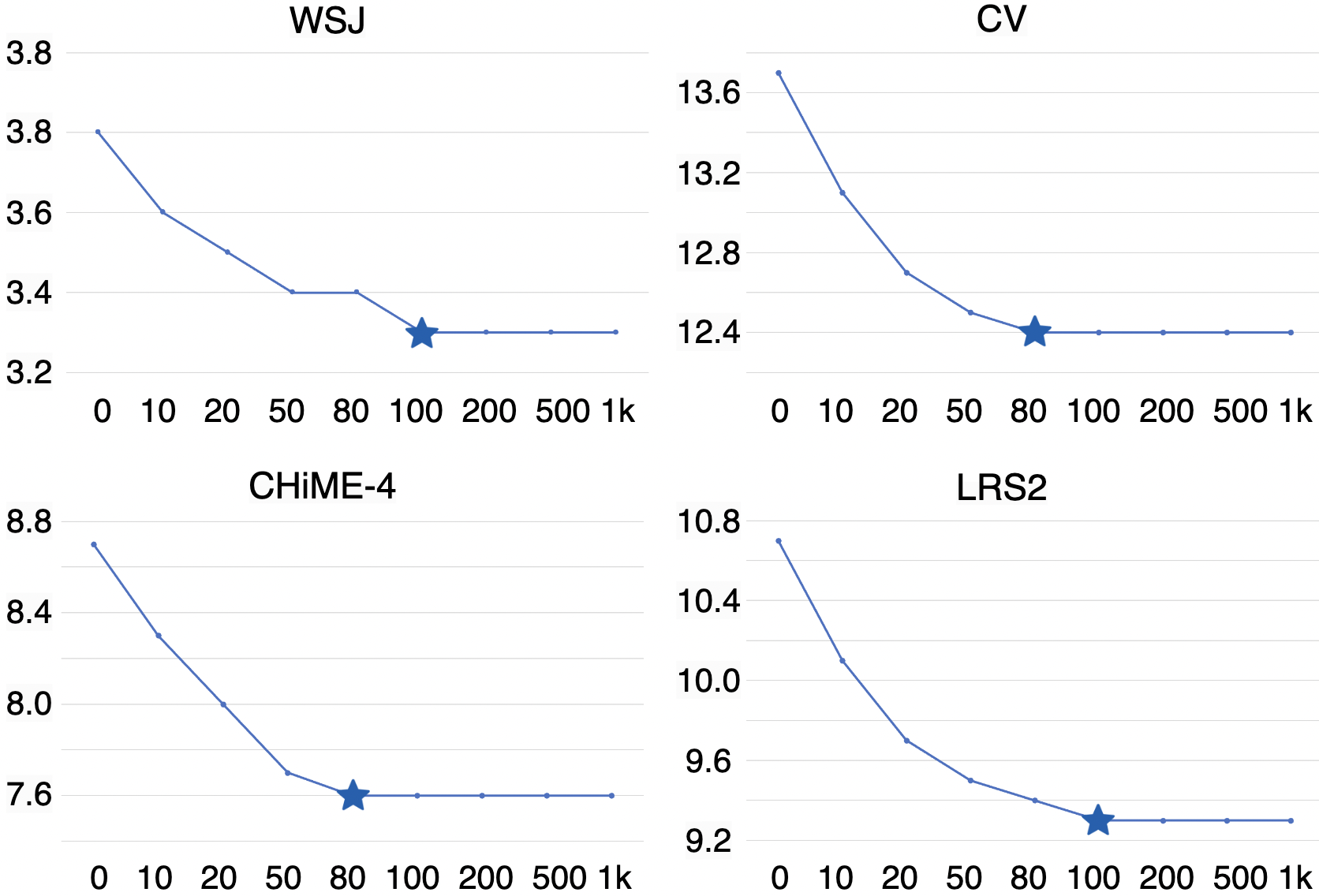}
\end{center}
\vspace{-0.2cm}
\caption{WER (\%) results of utilizing different numbers of validation samples for prior estimation.
The minimum required amount to obtain the best performance is highlighted in the star mark.
}
\vspace{-0.3cm}
\label{fig4}
\end{figure}

\vspace{0.1cm}
\noindent\noindent\textbf{LRS2}~\citep{chung2017lip}: Lip Reading Sentences 2 (LRS2) is a large-scale publicly available audio-visual dataset, consisting of 224 hours of video clips from BBC programs.
They randomly select 42,940 samples from its \emph{train} split as training set, and the rest of 2,259 samples are used for test set.

\vspace{0.1cm}
\noindent\textbf{ATIS}~\cite{hemphill1990atis}: Airline Travel Information System (ATIS) is a dataset comprising spoken queries for air travel information, including flight times, prices, and availability.
It contains 4,773 utterances recorded from over 500 speakers, which are separated into two parts to build training (3,964 samples) and test (809 samples) sets.

\vspace{0.1cm}
\noindent\textbf{CORAAL}~\cite{kendall2021corpus}: The Corpus of Regional African American Language (CORAAL) is the first public corpus of AAL speech data. 
It contains audio recordings along with the time-aligned orthographic transcriptions from over 150 sociolinguistic interviews.
To generate HyPoradise dataset, they select 1,728 samples as training set and 100 samples as test set.

\subsection{Validation Set Selection}
\label{assec:valid_set}
As mentioned in \S\ref{sssec:calibration}, our logits calibration method requires a validation set to calculate the prior distribution $\hat{\mathcal{P}}_\text{prior}(d_i)$.
To this end, we reserve a small portion of training samples to build the validation set.
To save the computation cost and time, we randomly select 100 samples from each ASR corpus in Table~\ref{table:statistics} for prior estimation~\cite{wu2021emotion}.
Relevant ablation study is illustrated in Fig.~\ref{fig4}, where we observe that around 100 validation samples are sufficient to estimate a reliable prior distribution for logits calibration on most datasets.

\section{Examples of Cloze test}
\label{asec:example_cloze}
Table~\ref{table:example_cloze} presents several examples of cloze test built from CHiME-4 \emph{test-real} set, where each example contains the context and several options.

\begin{figure}[t]
\begin{center}
\includegraphics[width=0.973\columnwidth]{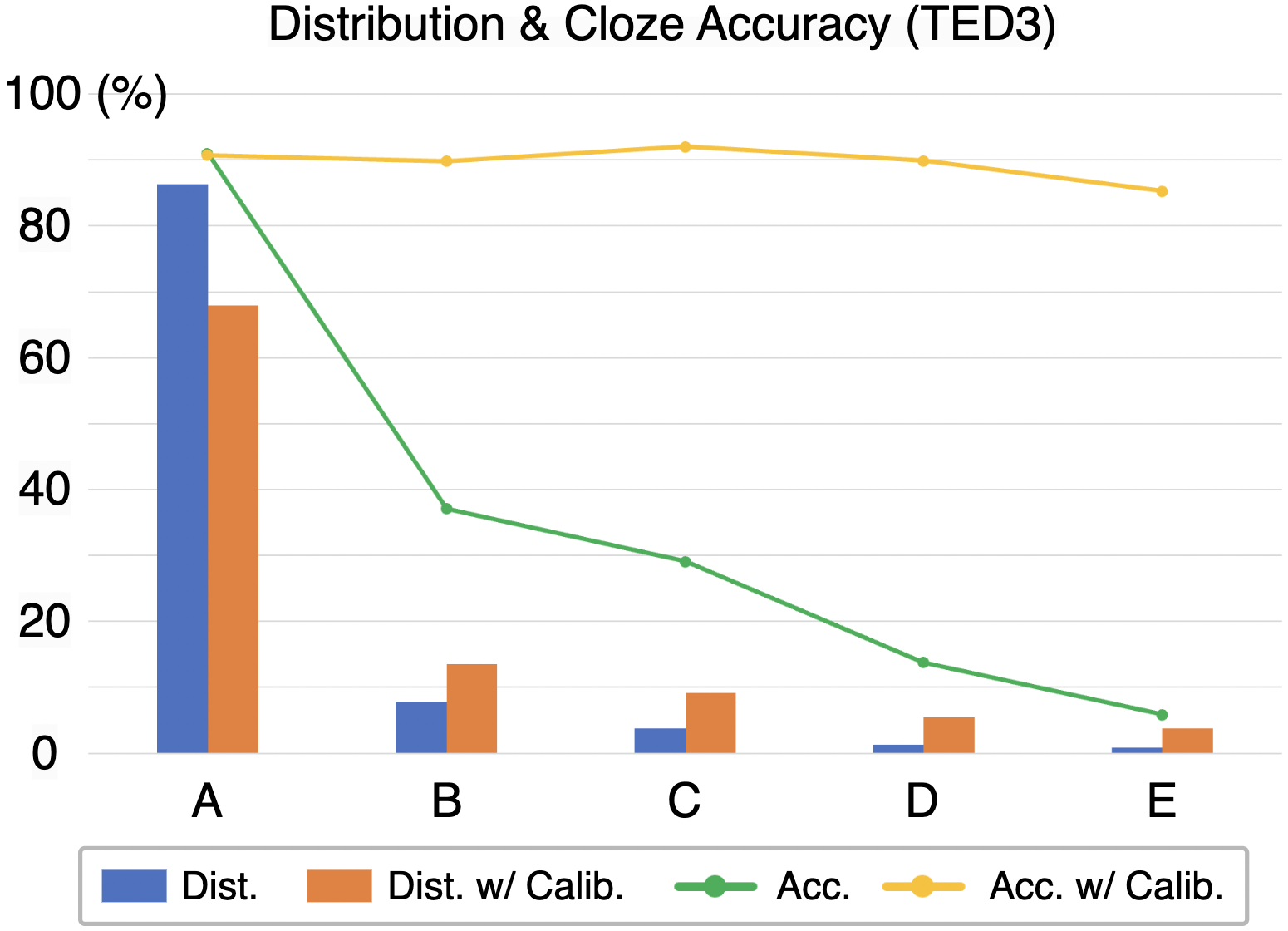}
\end{center}
\vspace{-0.2cm}
\caption{Distribution and cloze accuracy of five options with and without logits calibration on TED-LIUM 3 dataset.
The remarks follow that in Fig.~\ref{fig3}.
}
\label{fig5}
\end{figure}

\begin{table*}[t]
\caption{Examples of cloze test built from CHiME-4 \emph{test-real} set.
}
\label{table:example_cloze}
\centering
\resizebox{0.95\textwidth}{!}{
\begin{tabular}{c|l}
\toprule[1.2pt]
Example ID & F06\_443C0212\_CAF \\
\midrule
Cloze Context & yesterday is losers included \textit{\textcolor[HTML]{666666}{[Blank1]}} \\
\midrule
\multirow{6}{*}{Options} & \textit{\textcolor[HTML]{666666}{[Blank1]}}: \\
& A. automobiles \\
& B. all of you \\
& C. automobile \\
& D. all the ideas \\
& E. automakers \\
\midrule
Answer & A \\
\midrule[1.2pt]
Example ID & F06\_446C0204\_BUS \\
\midrule
Cloze Context & the consensus was that a new piece of paper is not required \textit{\textcolor[HTML]{666666}{[Blank1]}} one u s \textit{\textcolor[HTML]{666666}{[Blank2]}} \\
\midrule
\multirow{12}{*}{Options} & \textit{\textcolor[HTML]{666666}{[Blank1]}}: \\
& A. except \\
& B. said \\
& C. to be sent \\
& D. to set \\
& E. to send \\\cline{2-2}
& \textit{\textcolor[HTML]{666666}{[Blank2]}}: \\
& A. dollar \\
& B. diplomat \\
& C. dollar \\
& D. standard \\
& E. tip to them \\
\midrule
Answer & B\hspace{0.1cm} B \\
\midrule[1.2pt]
Example ID & M05\_440C020W\_STR \\
\midrule
Cloze Context & durable goods \textit{\textcolor[HTML]{666666}{[Blank1]}} frequently are highly volatile from month to month \\
\midrule
\multirow{6}{*}{Options} & \textit{\textcolor[HTML]{666666}{[Blank1]}}: \\
& A. and goods \\
& B. <NULL> \\
& C. and fluids \\
& D. and foods \\
& E. or goods \\
\midrule
Answer & A \\
\midrule[1.2pt]
Example ID & M05\_443C020R\_STR \\
\midrule
\multirow{2}{*}{Cloze Context} & as part of the marketing plan the company will begin airing television commercials \\
& during \textit{\textcolor[HTML]{666666}{[Blank1]}} on election night next tuesday \\
\midrule
\multirow{6}{*}{Options} & \textit{\textcolor[HTML]{666666}{[Blank1]}}: \\
& A. the prime time \\
& B. the fine time \\
& C. prime time \\
& D. fine time \\
& E. primetime \\
\midrule
Answer & C \\
\bottomrule[1.2pt]
\end{tabular}}
\vspace{-0.1cm}
\end{table*}

\end{document}